\documentclass[10pt,twocolumn,letterpaper]{article}

\usepackage{cvpr}              

\usepackage{graphicx}
\usepackage{amsmath}
\usepackage{amssymb}
\usepackage{booktabs}
\usepackage{multirow}
\usepackage{enumitem}
\usepackage[utf8]{inputenc}
\usepackage[T1]{fontenc}
\usepackage[english]{babel}
\usepackage[pagebackref,breaklinks,colorlinks]{hyperref}

\usepackage[capitalize]{cleveref}
\crefname{section}{Sec.}{Secs.}
\Crefname{section}{Section}{Sections}
\Crefname{table}{Table}{Tables}
\crefname{table}{Tab.}{Tabs.}

\def\cvprPaperID{25} 
\def\confName{UNCV-W}
\def\confYear{2024}

\begin{document}


\title{Evidential Transformers for Improved Image Retrieval}

\author{Danilo \DJ{}or\dj{}evi\'c\\
ETH Zürich\\
Rämistrasse 101\\
{\tt\small ddordevic@ethz.ch}
\and
Suryansh Kumar\\
Texas A\&M University, College Station\\
Visual Computing, College of PVFA\\
{\tt\small suryanshkumar@tamu.edu}
}
\maketitle

\begin{abstract}
  We introduce the Evidential Transformer, an uncertainty-driven transformer model for improved and robust image retrieval. In this paper, we make several contributions to content-based image retrieval (CBIR). We incorporate probabilistic methods into image retrieval,  achieving robust and reliable results, with evidential classification surpassing traditional training based on multiclass classification as a baseline for deep metric learning. Furthermore, we improve the state-of-the-art retrieval results on several datasets by leveraging the Global Context Vision Transformer (GC ViT) architecture. Our experimental results consistently demonstrate the reliability of our approach, setting a new benchmark in CBIR in all test settings on the Stanford Online Products (SOP) and CUB-200-2011 datasets.
\end{abstract}

\section{Introduction}
\label{sec:intro}


Content-based image retrieval is a well-known computer vision problem \cite{sivic2003video, csurka2004visual}. This problem aims to retrieve images from a database that are visually similar to a given query image. The similarity criterion used here is often measured between the vector representations of images. Such representations are generally sparse and relatively low-dimensional. They capture the semantic content of images by obtaining vector embeddings that contain local and global context information of images, and are vital to building an efficient image retrieval system \cite{aly_automatic_2009, cao_unifying_2020}.


Classical well-known pipelines for image retrieval tasks such as \cite{arandjelovic_three_2012, philbin2007object} begin by representing images using widely-adopted SIFT descriptors \cite{lowe_distinctive_2004} as image representations. Similarity between the query and database images is then assessed using metrics like cosine similarity, enabling the retrieval of visually similar images. Nevertheless, with the massive success of convolutional neural networks (CNNs) \cite{lecun1998gradient}, deep convolutional features have largely supplanted SIFT features due to their state-of-the-art results on several computer vision problems \cite{razavian_cnn_2014, cimpoi_deep_2015, babenko_aggregating_2015, radenovic2018fine, liuva, liu2023single, liu2024stereo, kaya2022uncertainty}.  Following these developments, CNN-based approaches such as \cite{babenko_aggregating_2015, babenko_neural_2014, mousavian_deep_2015} leverage neural codes for image retrieval\footnote{A neural code is a vector representation of an image, produced by the top layer of a CNN.}. Such neural codes perform competitively even when the code-producing network is trained on a task not directly related to image retrieval, like image classification \cite{zhai_classification_2019}. Prominent methods in this direction, such as \cite{radenovic_cnn_2016, radenovic_fine-tuning_2018}, fine-tune CNNs
for image retrieval without human annotation.  

Recently, Vision Transformer (ViT) architectures have outperformed CNNs across a variety of computer vision tasks \cite{dosovitskiy_image_2021, liu_swin_2021}.
In particular, methods using the ViT's outputs corresponding to the CLS-token as image descriptors for image retrieval have shown state-of-the-art results \cite{el-nouby_training_2021},
%
%
demonstrating that the ViT is capable of producing more informative embeddings, yielding superior results on multiple benchmark datasets, including CUB-200-2011 \cite{WahCUB_200_2011}, SOP \cite{song2016deep}, and InShop \cite{liuLQWTcvpr16DeepFashion}. Reranking methods that utilize attention have also emerged \cite{tan_instance-level_2022}. A recent extension, GC ViT \cite{hatamizadeh_global_2023}, proposed
a Vision Transformer architecture that efficiently models local and global contextual relations, producing a feature space where these relations induce a better separability between embeddings of different objects. 


However, current methods employ a generic similarity metric to rank retrieved images, which limits their ability to provide high-level retrieval information — 
specifically, the likelihood that retrieved images closely resemble the query image.
Consequently, we propose an evidential learning-driven approach to the transformer model for image retrieval. The idea is that the image similarity must not be confined to image class features but include other features such as the object's proximity to the camera, local and global scene context, number of pixel objects occupied in the query and database image, etc.
These aspects are challenging to model but significantly impact image retrieval effectiveness,
and thus, a probabilistic modeling approach can help account for this information in the GC ViT model.



Although there exist many approaches to probabilistic neural networks \cite{gawlikowski_survey_2022, gal_dropout_nodate, blundell_weight_2015, gal_bayesian_2016}, prior networks and the evidential learning paradigm reduce computational cost relative to ensemble approaches \cite{malinin_predictive_2018}. Therefore, we employ an evidential prior transformer instead of ensemble methods seen in \cite{cinquin_bayesian_2021, sankararaman_bayesformer_2022, gal_bayesian_2016}, leveraging evidential deep learning \cite{sensoy_evidential_2018, joo_being_2020, malinin_predictive_2018, mena_dirichlet_2019}. Motivated by recent works demonstrating that classification is a strong baseline for deep metric learning \cite{zhai_classification_2019}, we study the potential of deep evidential classification in providing a more informed baseline for deep metric learning. By setting an evidential prior over the predicted softmax probabilities, we infer image embedding in probabilistic terms.


Subsequently, in this paper, we introduce an approach that efficiently combines attention-based feature maps with uncertainty quantification methods to improve the image-retrieval performance and robustness. Concretely, how embedding uncertainties can be included in the image retrieval process is presented. We observed that uncertainty-driven transformer-based neural representations outperform existing methods in content-based category-level image retrieval. 
%
%
%
%
To summarize, our key \textbf{contributions} are:

\begin{itemize}[noitemsep, rightmargin=1pt, topsep=0pt]
    \item We show that deep evidential classification is a strong baseline for deep metric learning by achieving state-of-the-art results on the CUB-200-2011 dataset.
    
    \item We present a novel, task-agnostic re-ranking method based on uncertainty estimates, which outperforms baseline methods that do not incorporate re-ranking.

    \item Our work shows that Dirichlet probability distribution parameters are good neural codes for image retrieval. Additionally, this work introduce a novel continuous embedding method, where each image is mapped into a distribution space and subsequently compared using the Bhattacharyya distance.
    
    \item Our rigorous experiments demonstrate that the our uncertainty-driven tranformer approach significantly outperforms current state-of-the-art image retrieval methods on the CUB-200-2011 and SOP datasets.
\end{itemize}


\section{Methodology}
Evidential learning is an effective method for quantifying uncertainty in image retrieval tasks. In contrast to traditional deterministic neural networks that output point predictions,
evidential learning explicitly models uncertainty, providing a more robust and informative framework for applications in image retrieval.

The main advantage of evidential networks is that they do not produce a soft-max categorical probability distribution over the class labels but rather a second-order distribution over the soft-max probabilities, allowing them to reason about the uncertainty of the second-order distribution. Such uncertainty can arise from out-of-distribution samples in the dataset, or from images that have embeddings similar to the query but belong to a different class. Assigning a lower rank to these examples could improve the quality of retrieval. This is particularly useful in scenarios where it is crucial to assess the confidence of the model's predictions, such as in the retrieval of images from large and diverse datasets.


In evidential learning, the output of a neural network is interpreted as a $K$-dimensional parameter vector $\boldsymbol{\alpha} = \left[ \alpha_1, \dots, \alpha_K\right]$ of a Dirichlet distribution, which provides a second-order probability distribution over softmax class probabilities $\mathbf{p}$: 
    \begin{equation}
        D(\mathbf{p} \mid \boldsymbol{\alpha})= \begin{cases}\frac{1}{B(\boldsymbol{\alpha})} \prod_{i=1}^K p_i^{\alpha_i-1} & \text { for } \mathbf{p} \in \mathcal{S}_K \\ 0 & \text { otherwise }\end{cases}
    \end{equation}

where $K$ is the number of classes in a classification task, and $\mathcal{S}_K$ is an $K-1$ simplex:
    \begin{equation}
        \mathcal{S}_K=\left\{\mathbf{p} \mid \sum_{i=1}^K p_i=1 \text { and } 0 \leq p_1, \ldots, p_K \leq 1\right\}    
    \end{equation} \label{eq:simplex-def}
    
This is a significant departure from the softmax prediction, which only offers a single set of class probabilities $\mathbf{p}$ without any indication of uncertainty.
The parameters \(\alpha_k\) are derived from the evidence \(e_k\) for each class $k \in \{1,2, \dots, K\}$, such that \(\alpha_k = e_k + 1\).
The belief mass \cite{Dempster2008} \(b_k\) for each class and the overall uncertainty \(u\) are computed as follows:
\begin{equation}
    b_k = \frac{e_k}{S};~~~~u = \frac{K}{S} 
\end{equation}
%
%
where \(S = \sum_{i=1}^K (e_i + 1)\). This formulation ensures that the uncertainty \(u\) is inversely proportional to the total evidence \(S\). When no evidence in support of a prediction is available, the model gives maximum uncertainty (\(u = 1\)).

The learning process involves training a neural network to predict the parameters of the Dirichlet distribution $\boldsymbol{\alpha}$, which are parameterized by the network weights $\Theta$. An evidential network is trained on a classification task by optimizing Bayes risk function that incorporates both prediction accuracy and its uncertainty. This approach is based on empirical findings from \cite{sensoy_evidential_2018}:
\begin{align}
        \mathcal{L}_i(\Theta) &=\int\left\|\mathbf{y}_i-\mathbf{p}_i\right\|_2^2 \frac{1}{B\left(\boldsymbol{\alpha}_i\right)} \prod_{i=1}^K p_{i j}^{\alpha_{i j}-1} d \mathbf{p}_i \\
        &=\sum_{j=1}^K \mathbb{E}\left[y_{i j}^2-2 y_{i j} p_{i j}+p_{i j}^2\right] \\
        &=\sum_{j=1}^K\left(y_{i j}^2-2 y_{i j} \mathbb{E}\left[p_{i j}\right]+\mathbb{E}\left[p_{i j}^2\right]\right) \label{eq: uncertainty_mse}
    \end{align}

The loss function is designed to minimize the prediction error while encouraging the model to produce low-variance distributions. We explore four possible ways of embedding Dirichlet uncertainties into the category-level image retrieval framework, described in the following sections.


\subsection{Evidential Classification as a Deep Metric Learning Baseline}
Building on the methodology presented in \cite{sensoy_evidential_2018}, we train DeiT-S \cite{avidan_deit_2022} and GC ViT Tiny models for the evidential classification task using the CUB-200-2011 dataset. This approach reinterprets the networks' outputs and introduces a novel loss function, described in \autoref{eq: uncertainty_mse}. By leveraging the CLS token from the Vision Transformers as an image embedding, we perform image retrieval by comparing the embeddings of the query image with those of the database images using cosine similarity. The retrieval process is illustrated in \autoref{fig:retrieval-system}.

\subsection{$\boldsymbol{\alpha}$-embeddings: Using Dirichlet Distribution Parameters as Neural Codes}

\begin{figure}
        \centering
        \includegraphics[width=0.8\linewidth]{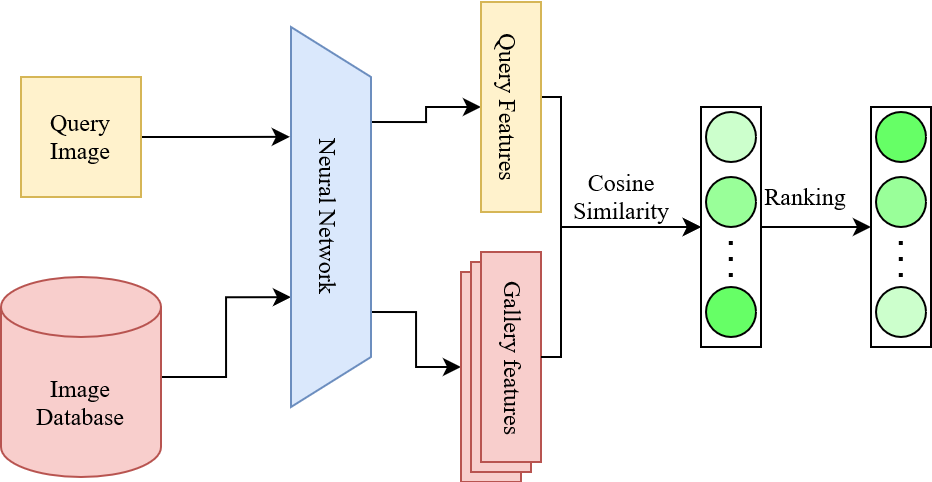}
        \caption{\textit{Retrieval based on image embeddings}. CLS token of the vision transformer (or $\alpha$ vector) is taken as the image embedding, and is used for matching between image pairs using cosine similarity.}
        \label{fig:retrieval-system}
    \end{figure}

In this experiment, instead of using the CLS Token, we employ the final-layer outputs of the network, which now represent the parameters $\boldsymbol{\alpha}$ of the Dirichlet distribution rather than the softmax probability distribution $\mathbf{p}$. These $\boldsymbol{\alpha}$ parameters are utilized directly as image embeddings and are compared using the $L_2$ distance metric. This approach aims to evaluate the effectiveness of $\boldsymbol{\alpha}$ parameters as neural representations. The diagram in \autoref{fig:retrieval-system} also applies to this retrieval paradigm.

\subsection{Uncertainty-Driven Retrieval Reranking}
\begin{figure}
        \centering
        \includegraphics[width=\linewidth]{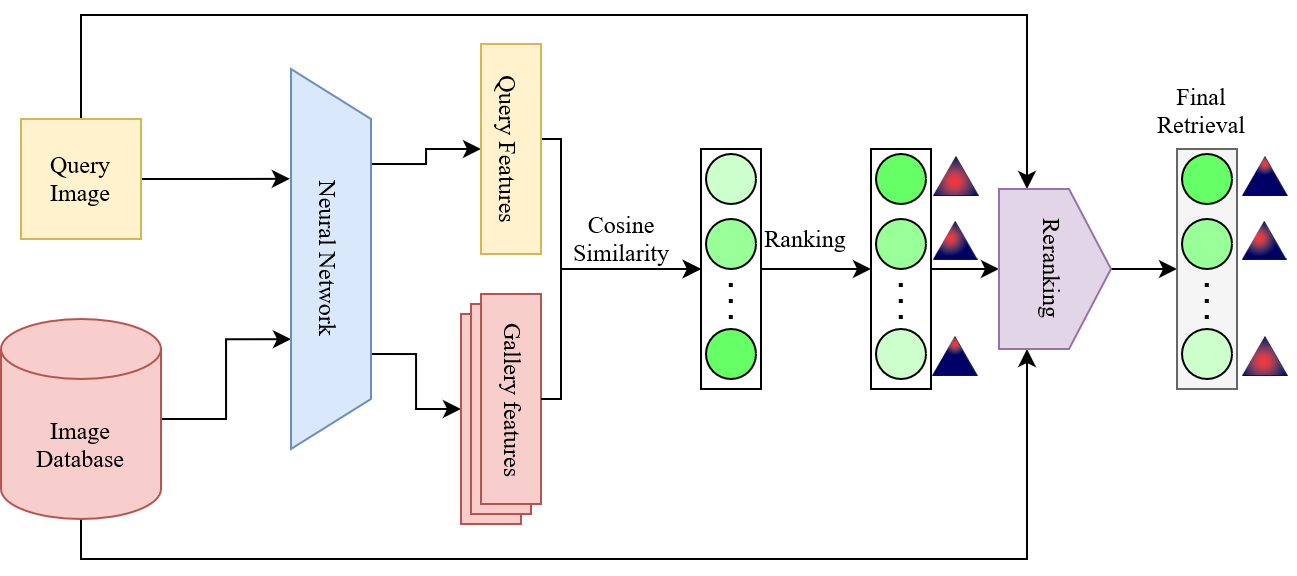}
        \caption{\textit{Uncertainty-based re-ranking}: First retrieval is performed the regular way, by comparing the image descriptors using the cosine similarity. However, after that, a separate, evidential neural network computes uncertainties for top K results, after which re-ranking based on these uncertainties is performed.}
        \label{fig:retrieval-rerank}
    \end{figure}

In this setup, we utilize a contrastively-trained GC ViT model for feature extraction. Subsequently, an evidential GC ViT is employed to derive uncertainty values for each of the top $N$ results. These results are then reranked in ascending order of uncertainty. The process is depicted in \autoref{fig:retrieval-rerank}.

\subsection{Distribution-Based Distances}

\begin{figure}
        \centering
        \includegraphics[width=0.9\linewidth]{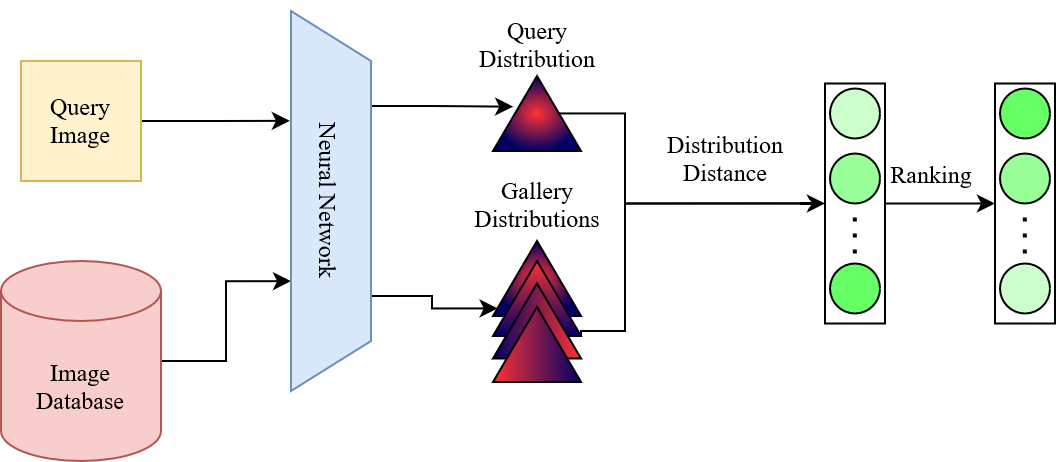}
        \caption{\textit{Retrieval based on continuous distribution embeddings}. Images are mapped to $\alpha$-parameterized Dirichlet distributions, after which matching is performed based on negative Bhattacharyya distance between them, instead of the inner product.}
        \label{fig:retrieval-bhat}
    \end{figure}

Given that the network outputs now represent the parameters of the Dirichlet distribution, which encode the uncertainty of the network's predictions, this approach explores the feasibility of directly embedding images into distributions. Instead of comparing vector embeddings, we directly compare these distributional embeddings using the Bhattacharyya distance \cite{rauber_bhattacharyya_2008}. Images that are similar are close in distribution space, indicated by a small Bhattacharyya distance between them. The method is illustrated in \autoref{fig:retrieval-bhat}.


\section{Experiements and Results}
\begin{figure}
        \centering
        \includegraphics[width=0.85\linewidth]{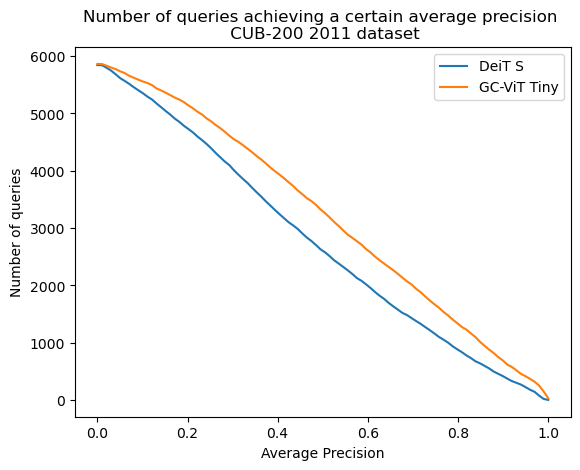}
        \caption{Number of queries exceeding a given average precision on the CUB-200-2011 dataset. GC ViT outperforms the DeiT across all average precision thresholds.}
        \label{fig:curves-cub}
\end{figure}

\begin{figure}
        \centering
        \includegraphics[width=0.85\linewidth]{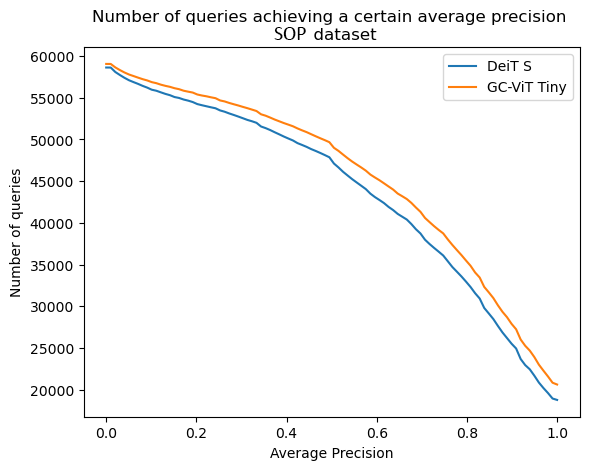}
        \caption{Number of queries exceeding a given average precision on the SOP dataset. GC ViT outperforms the DeiT across all average precision thresholds.}
        \label{fig:curves-sop}
\end{figure}

First we perform experiments which concern the choice of the best-performing backbone architecture for embedding the images for further use in retrieval. Following \cite{el-nouby_training_2021}, all models used in this paper are fine-tuned either on a classification task or on the metric-learning task using entropy-regularization. Global Context Vision Transformer has shown to outperform the DeiT-S used in \cite{el-nouby_training_2021}, on both the CUB-200-2011 and SOP datasets, by a wide margin, as demonstrated in \autoref{tab:deit_vs_gcvit_sop_cub}, \autoref{fig:curves-cub}, \autoref{fig:curves-sop}.

This led us to adopt the GC ViT as the backbone architecture, and conduct further experiments on the CUB-200-2011 dataset to explore the effectiveness of proposed probabilistic approaches. Results are summarized in \autoref{tab:cub200-uncertainty-results}. Evidential classification excelled, surpassing regular classification in terms of the Recall@K metric. The best performing method was Uncertainty-Driven Reranking, while $\boldsymbol{\alpha}$-embeddings and Distribution-embedding methods fell short, yielding results below the baseline.

\begin{table}
        \centering
        \caption{Comparison between the contrastively-trained DeiT-S and GC ViT-T models in category-level image retrieval tasks on the Stanford Online Products and CUB-200-2011 datasets.}
        \begin{tabular}{@{}ll|ll@{}}
             \toprule
             \multicolumn{4}{c}{Recall@K [\%]} \\ \cmidrule{3-4}
             Dataset & K & DeiT-S & GC ViT-T \\
             \midrule
             \multirow{4}{*}{SOP} & 1 & 83.28 & \textbf{85.45} \ [+2.17\%] \\
             & 10 & 93.08 & \textbf{94.38} \ [+1.30\%] \\
             & 100 & 96.86 & \textbf{97.60} \ [+0.74\%] \\
             & 1000 & 98.86 & \textbf{99.16} \ [+0.30\%] \\
             \midrule
             \multirow{4}{*}{CUB} & 1 & 62.64 & \textbf{70.80} \ [+13.02\%] \\
             & 2 & 73.21 & \textbf{79.54} \ [+8.64\%] \\
             & 4 & 81.87 & \textbf{86.85} \ [+6.08\%] \\
             & 8 & 88.47 & \textbf{92.40} \ [+4.44\%] \\
             \bottomrule
        \end{tabular}
\label{tab:deit_vs_gcvit_sop_cub}
\end{table}

\begin{figure}
        \centering
        \includegraphics[width=1.0\linewidth]{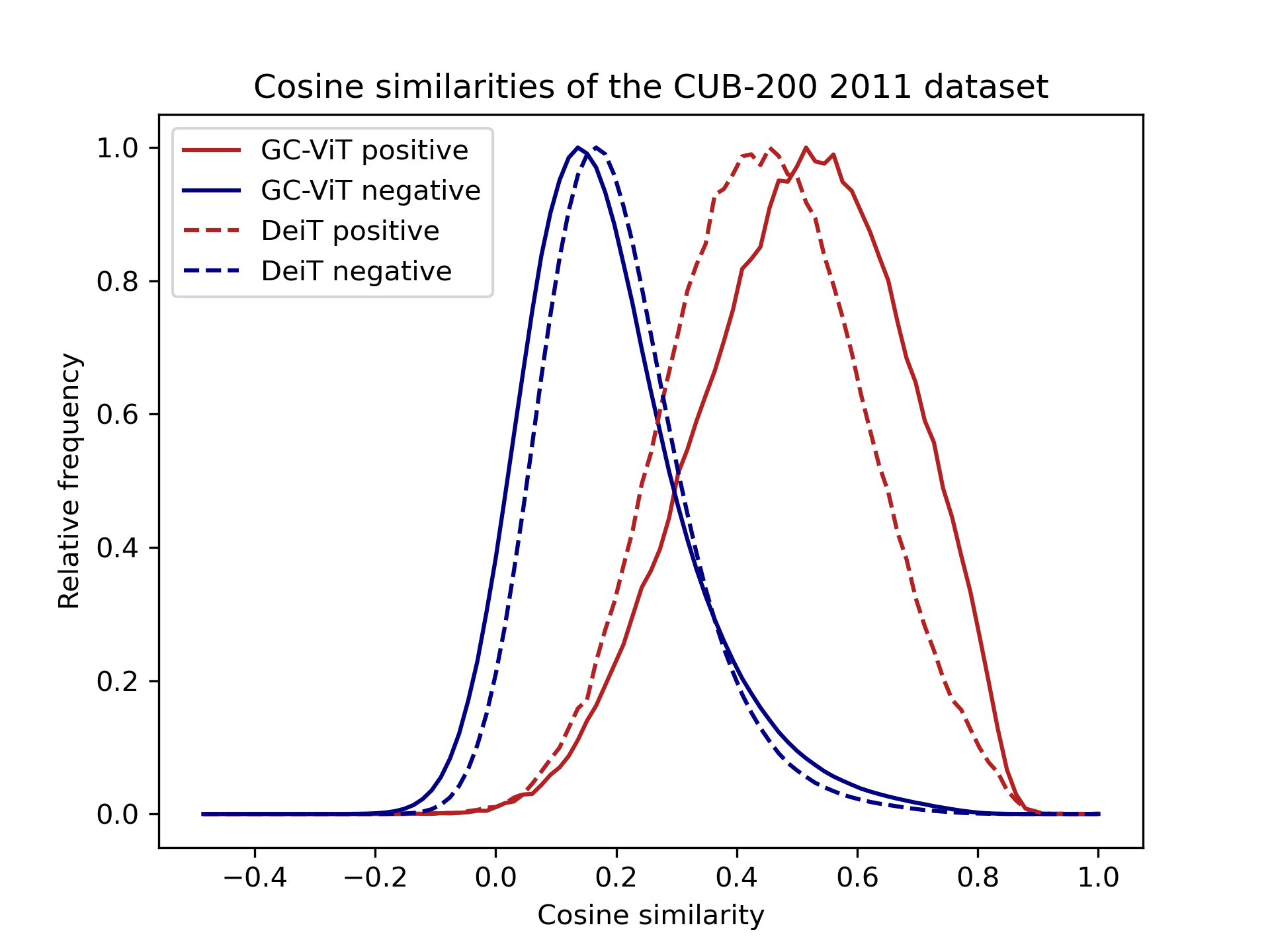}
        \caption{Cosine similiarities of positive and negative pairs produced by two architectures. Greater separation observed with the GC ViT indicates better processing of both local and global relations within images.}
        \label{fig:cosine-hist-cub}
\end{figure}

\begin{table}
        \centering
        \caption{\textbf{Evidential retrieval.} Recall@K on the CUB-200 2011 dataset, achieved with the GC ViT Tiny model, trained under the various evidential network frameworks. Each row corresponds to a different method of retrieval.}
        \begin{tabular}{l|llll}
             \toprule
             & \multicolumn{4}{c}{Recall@K [\%]} \\
             \cmidrule{2-5} 
             \textbf{Retrieval type }/ \textbf{K} & 1 & 2 & 4 & 8 \\
             \midrule
             Classification (base) & 72.89 & 82.65 & 88.47 & 92.94 \\
             Evidential class. & 80.22 & \textbf{86.38} & 90.23 & 93.15 \\
             Uncertainty rerank. & \textbf{80.79} & 86.09 & \textbf{90.42} & \textbf{93.17} \\
             $\boldsymbol{\alpha}$-embeddings & 53.87 & 66.26 & 77.08 & 85.04 \\
             Distribution emb. & 4.00 & 4.12 & 5.33 & 5.98 \\
             \bottomrule
        \end{tabular}
\label{tab:cub200-uncertainty-results}
\end{table}

\section{Conclusion}
Our research advances the field of content-based image retrieval by introducing and leveraging evidential uncertainty quantification methods. Specifically, we utilize evidential learning to enhance general image retrieval quality, we introduce a general task-agnostic reranking scheme based on uncertainty values, benchmarks novel transformer architectures, and establish evidential classification as a good baseline for metric learning. Our findings suggest that incorporating uncertainty estimates into image retrieval systems can substantially improve their quality, reliability and interpretability. Future research could further explore adversarial robustness, different distribution-based embeddings as well as other probabilistic methods to build upon the foundational steps we established.

{\small
\bibliographystyle{ieee_fullname}
\bibliography{egbib}

\begin{thebibliography}{10}\itemsep=-1pt

\bibitem{aly_automatic_2009}
Mohamed Aly, Peter Welinder, Mario Munich, and Pietro Perona.
\newblock Automatic discovery of image families: Global vs. local features.
\newblock In {\em 2009 16th {IEEE} International Conference on Image Processing ({ICIP})}, pages 777--780. {IEEE}, 2009.

\bibitem{arandjelovic_three_2012}
Relja Arandjelovi{\'c} and Andrew Zisserman.
\newblock Three things everyone should know to improve object retrieval.
\newblock {\em 2012 IEEE Conference on Computer Vision and Pattern Recognition}, pages 2911--2918, 2012.

\bibitem{babenko_aggregating_2015}
Artem Babenko and Victor Lempitsky.
\newblock Aggregating deep convolutional features for image retrieval.

\bibitem{babenko_neural_2014}
Artem Babenko, Anton Slesarev, Alexandr Chigorin, and Victor Lempitsky.
\newblock Neural codes for image retrieval.

\bibitem{blundell_weight_2015}
Charles Blundell, Julien Cornebise, Koray Kavukcuoglu, and Daan Wierstra.
\newblock Weight {Uncertainty} in {Neural} {Networks}, May 2015.
\newblock arXiv:1505.05424 [cs, stat].

\bibitem{cao_unifying_2020}
Bingyi Cao, Andre Araujo, and Jack Sim.
\newblock Unifying deep local and global features for image search.

\bibitem{cimpoi_deep_2015}
Mircea Cimpoi, Subhransu Maji, Iasonas Kokkinos, and Andrea Vedaldi.
\newblock Deep filter banks for texture recognition, description, and segmentation.

\bibitem{cinquin_bayesian_2021}
Tristan Cinquin.
\newblock The bayesian transformer.
\newblock Master thesis, ETH Zurich, Zurich, 2021.

\bibitem{csurka2004visual}
Gabriella Csurka, Christopher~R. Dance, Lixin Fan, Jutta~Katharina Willamowski, and C{\'e}dric Bray.
\newblock Visual categorization with bags of keypoints.
\newblock In {\em European Conference on Computer Vision}, 2002.

\bibitem{Dempster2008}
Arthur~P. Dempster.
\newblock {\em A Generalization of Bayesian Inference}, pages 73--104.
\newblock Springer Berlin Heidelberg, Berlin, Heidelberg, 2008.

\bibitem{dosovitskiy_image_2021}
Alexey Dosovitskiy, Lucas Beyer, Alexander Kolesnikov, Dirk Weissenborn, Xiaohua Zhai, Thomas Unterthiner, Mostafa Dehghani, Matthias Minderer, Georg Heigold, Sylvain Gelly, Jakob Uszkoreit, and Neil Houlsby.
\newblock An image is worth 16x16 words: Transformers for image recognition at scale.

\bibitem{el-nouby_training_2021}
Alaaeldin El-Nouby, Natalia Neverova, Ivan Laptev, and Hervé Jégou.
\newblock Training vision transformers for image retrieval.

\bibitem{gal_bayesian_2016}
Yarin Gal and Zoubin Ghahramani.
\newblock Bayesian convolutional neural networks with bernoulli approximate variational inference.

\bibitem{gal_dropout_nodate}
Yarin Gal and Zoubin Ghahramani.
\newblock Dropout as a bayesian approximation: Representing model uncertainty in deep learning.
\newblock In Maria~Florina Balcan and Kilian~Q. Weinberger, editors, {\em Proceedings of The 33rd International Conference on Machine Learning}, volume~48 of {\em Proceedings of Machine Learning Research}, pages 1050--1059, New York, New York, USA, 20--22 Jun 2016. PMLR.

\bibitem{gawlikowski_survey_2022}
Jakob Gawlikowski, Cedrique Rovile~Njieutcheu Tassi, Mohsin Ali, Jongseok Lee, Matthias Humt, Jianxiang Feng, Anna Kruspe, Rudolph Triebel, Peter Jung, Ribana Roscher, Muhammad Shahzad, Wen Yang, Richard Bamler, and Xiao~Xiang Zhu.
\newblock A survey of uncertainty in deep neural networks.

\bibitem{hatamizadeh_global_2023}
Ali Hatamizadeh, Hongxu Yin, Greg Heinrich, Jan Kautz, and Pavlo Molchanov.
\newblock Global context vision transformers.

\bibitem{joo_being_2020}
Taejong Joo, Uijung Chung, and Min-Gwan Seo.
\newblock Being bayesian about categorical probability.

\bibitem{kaya2022uncertainty}
Berk Kaya, Suryansh Kumar, Carlos Oliveira, Vittorio Ferrari, and Luc Van~Gool.
\newblock Uncertainty-aware deep multi-view photometric stereo.
\newblock In {\em Proceedings of the IEEE/CVF Conference on Computer Vision and Pattern Recognition}, pages 12601--12611, 2022.

\bibitem{lecun1998gradient}
Y. Lecun, L. Bottou, Y. Bengio, and P. Haffner.
\newblock Gradient-based learning applied to document recognition.
\newblock {\em Proceedings of the IEEE}, 86(11):2278--2324, 1998.

\bibitem{liu2023single}
Ce Liu, Suryansh Kumar, Shuhang Gu, Radu Timofte, and Luc Van~Gool.
\newblock Single image depth prediction made better: A multivariate gaussian take.
\newblock In {\em Proceedings of the IEEE/CVF Conference on Computer Vision and Pattern Recognition}, pages 17346--17356, 2023.

\bibitem{liuva}
Ce Liu, Suryansh Kumar, Shuhang Gu, Radu Timofte, and Luc Van~Gool.
\newblock Va-depthnet: A variational approach to single image depth prediction.
\newblock In {\em The Eleventh International Conference on Learning Representations}, 2023.

\bibitem{liu2024stereo}
Ce Liu, Suryansh Kumar, Shuhang Gu, Radu Timofte, Yao Yao, and Luc Van~Gool.
\newblock Stereo risk: A continuous modeling approach to stereo matching.
\newblock {\em arXiv preprint arXiv:2407.03152}, 2024.

\bibitem{liu_swin_2021}
Ze Liu, Yutong Lin, Yue Cao, Han Hu, Yixuan Wei, Zheng Zhang, Stephen Lin, and Baining Guo.
\newblock Swin transformer: Hierarchical vision transformer using shifted windows.

\bibitem{liuLQWTcvpr16DeepFashion}
Ziwei Liu, Ping Luo, Shi Qiu, Xiaogang Wang, and Xiaoou Tang.
\newblock Deepfashion: Powering robust clothes recognition and retrieval with rich annotations.
\newblock In {\em Proceedings of IEEE Conference on Computer Vision and Pattern Recognition (CVPR)}, June 2016.

\bibitem{lowe_distinctive_2004}
David~G. Lowe.
\newblock Distinctive image features from scale-invariant keypoints.
\newblock {\em International Journal of Computer Vision}, 60(2):91--110, 2004.

\bibitem{malinin_predictive_2018}
Andrey Malinin and Mark Gales.
\newblock Predictive uncertainty estimation via prior networks.

\bibitem{mena_dirichlet_2019}
José Mena, Oriol Pujol, and Jordi Vitrià.
\newblock Dirichlet uncertainty wrappers for actionable algorithm accuracy accountability and auditability, Dec. 2019.
\newblock arXiv:1912.12628 [cs, stat].

\bibitem{mousavian_deep_2015}
Arsalan Mousavian and Jana Kosecka.
\newblock Deep convolutional features for image based retrieval and scene categorization.

\bibitem{philbin2007object}
James Philbin, Ondrej Chum, Michael Isard, Josef Sivic, and Andrew Zisserman.
\newblock Object retrieval with large vocabularies and fast spatial matching.
\newblock In {\em 2007 IEEE conference on computer vision and pattern recognition}, pages 1--8. IEEE, 2007.

\bibitem{radenovic2018fine}
Filip Radenovi{\'c}, Giorgos Tolias, and Ond{\v{r}}ej Chum.
\newblock Fine-tuning cnn image retrieval with no human annotation.
\newblock {\em IEEE transactions on pattern analysis and machine intelligence}, 41(7):1655--1668, 2018.

\bibitem{radenovic_cnn_2016}
Filip Radenović, Giorgos Tolias, and Ondřej Chum.
\newblock {CNN} image retrieval learns from {BoW}: Unsupervised fine-tuning with hard examples.

\bibitem{radenovic_fine-tuning_2018}
Filip Radenović, Giorgos Tolias, and Ondřej Chum.
\newblock Fine-tuning {CNN} image retrieval with no human annotation.

\bibitem{rauber_bhattacharyya_2008}
T.~W. Rauber, A. Conci, T. Braun, and K. Berns.
\newblock Bhattacharyya probabilistic distance of the {Dirichlet} density and its application to {Split}-and-{Merge} image segmentation.
\newblock In {\em 2008 15th {International} {Conference} on {Systems}, {Signals} and {Image} {Processing}}, pages 145--148, Bratislava, Slovakia, June 2008. IEEE.

\bibitem{razavian_cnn_2014}
Ali~Sharif Razavian, Hossein Azizpour, Josephine Sullivan, and Stefan Carlsson.
\newblock {CNN} features off-the-shelf: an astounding baseline for recognition.

\bibitem{sankararaman_bayesformer_2022}
Karthik~Abinav Sankararaman, Sinong Wang, and Han Fang.
\newblock {BayesFormer}: Transformer with uncertainty estimation.

\bibitem{sensoy_evidential_2018}
Murat Sensoy, Lance Kaplan, and Melih Kandemir.
\newblock Evidential deep learning to quantify classification uncertainty.

\bibitem{sivic2003video}
Sivic and Zisserman.
\newblock Video google: A text retrieval approach to object matching in videos.
\newblock In {\em Proceedings ninth IEEE international conference on computer vision}, pages 1470--1477. IEEE, 2003.

\bibitem{song2016deep}
Hyun~Oh Song, Yu Xiang, Stefanie Jegelka, and Silvio Savarese.
\newblock Deep metric learning via lifted structured feature embedding.
\newblock In {\em IEEE Conference on Computer Vision and Pattern Recognition (CVPR)}, 2016.

\bibitem{tan_instance-level_2022}
Fuwen Tan, Jiangbo Yuan, and Vicente Ordonez.
\newblock Instance-level image retrieval using reranking transformers.

\bibitem{avidan_deit_2022}
Hugo Touvron, Matthieu Cord, and Herv\'{e} J\'{e}gou.
\newblock Deit iii: Revenge of the vit.
\newblock In {\em Computer Vision – ECCV 2022: 17th European Conference, Tel Aviv, Israel, October 23–27, 2022, Proceedings, Part XXIV}, page 516–533, Berlin, Heidelberg, 2022. Springer-Verlag.

\bibitem{WahCUB_200_2011}
C. Wah, S. Branson, P. Welinder, P. Perona, and S. Belongie.
\newblock The {Caltech}-{UCSD} {Birds}-200-2011 {Dataset}.
\newblock Technical Report CNS-TR-2011-001, California Institute of Technology, 2011.

\bibitem{zhai_classification_2019}
Andrew Zhai and Hao-Yu Wu.
\newblock Classification is a {Strong} {Baseline} for {Deep} {Metric} {Learning}, Aug. 2019.
\newblock arXiv:1811.12649 [cs].

\end{thebibliography}
}

\end{document}